\documentclass[final]{hld2026} % Author names withheld

\usepackage{hyperref}       % hyperlinks
\usepackage{url}            % simple URL typesetting
\usepackage{booktabs}       % professional-quality tables
\usepackage{amsfonts}       % blackboard math symbols
\usepackage{nicefrac}       % compact symbols for 1/2, etc.
\usepackage{microtype}      % microtypography
\usepackage{xcolor}         % colors
\usepackage{algorithm,algorithmic}
\usepackage{amsmath}
\usepackage{mathtools}
\usepackage{caption}
\usepackage{tcolorbox}
\usepackage{subcaption}
\usepackage{multirow}
\mathtoolsset{showonlyrefs=true}

\title[Rethinking Bregman Divergences in Kronecker-Factored Optimizers]{Rethinking Bregman Divergences in Kronecker-Factored Optimizers}

\hldauthor{%
\Name{Bing Liu}$^*$ \Email{bing\_liu@zju.edu.cn}\\
\addr {College of Control Science and Engineering, Zhejiang University} 
\AND
\Name{Wenjie Zhou}$^*$ \Email{zj4323005@gmail.com}\\
\addr {State Key Laboratory of AI Safety, Institute of Computing Technology, Chinese Academy of Sciences}
\AND
\Name{Chengcheng Zhao}$^{\dagger}$ \Email{chengchengzhao@zju.edu.cn}\\
\addr {College of Control Science and Engineering, Zhejiang University} 
\AND
\textnormal{$^*$ Equal contribution\qquad $^{\dagger}$ Corresponding author}
}
\begin{document}

\maketitle

\begin{abstract}
Shampoo-style optimizers approximate gradient covariance matrices using Kronecker-factored structures. Recent work~\cite{lin2026understanding} showed that such approximations can be viewed as projections under Bregman matrix divergences, leading to different Kronecker-factored preconditioners. However, it remains unclear what role the choice of divergence plays when the covariance is not exactly Kronecker-factored. We study this question through the spectrum of the covariance matrix. We show that Frobenius, von Neumann, and LogDet divergences distribute the unavoidable Kronecker approximation error differently across the covariance spectrum. We further show that their Kronecker factors are governed by divergence-weighted residuals rather than the raw approximation error, explaining how these spectral preferences are realized in the resulting preconditioners. Empirically, we observe that the top covariance eigenspace is substantially better aligned with the Hessian matrix, while the tail spectrum is much noisier and unreliable. Motivated by these findings, we propose a subspace-aware Kronecker optimizer that applies eigenvalue-based preconditioning in the top subspace and uses an adaptive isotropic acceleration constant in the bottom subspace. 
\end{abstract}

%\begin{keywords}%
%  List of keywords%
%\end{keywords}

\section{Introduction}

Shampoo-style optimizers~\cite{gupta2018shampoo} exploit the matrix structure of gradients by replacing dense second-order preconditioners with Kronecker-factored approximations. For a matrix-valued parameter with gradient $G$, this corresponds to approximating the gradient covariance matrix $C=\mathbb{E}[\mathrm{vec}(G)\mathrm{vec}(G)^\top]$ by a product $L\otimes R$. Lin et al.~\cite{lin2026understanding} showed that several bilateral Shampoo variants arise from minimizing Bregman matrix divergences between $C$ and $L\otimes R$. However, this unification does not explain what happens when the covariance is not exactly Kronecker-factored. For a generic dense covariance matrix, nonzero approximation error is unavoidable, so different divergences emphasize different parts of the spectrum. Furthermore, since the covariance significantly differs from the Hessian, meaning that not all its spectrum is suitable for preconditioning~\cite{kunstner2019limitations,thomas2020interplay}. We therefore ask: 

\vspace{-0.4em}
\begin{tcolorbox}[
    colback=gray!15,
    colframe=gray!15,
    boxrule=0.4pt,
    arc=1.5pt,
    left=4pt,
    right=4pt,
    top=3pt,
    bottom=3pt,
    width=\linewidth
]
\centering
\emph{Q: When exact Kronecker matching is impossible, which parts of the covariance spectrum should a structured optimizer approximate and trust?}
\end{tcolorbox}
\vspace{-0.4em}

Our work connects two issues: the approximation geometry induced by different Bregman matrix divergences, and the spectral reliability of covariance-based preconditioning. We make this connection through three contributions below:
\begin{itemize}
    \item \textbf{Spectral geometry of Bregman Kronecker approximation.} We characterize the eigenvalue-dependent penalties induced by Frobenius, von Neumann, and LogDet divergences in covariance approximation, and show that the resulting bilateral preconditioners reflect these spectral preferences through divergence-weighted residuals on the Kronecker manifold.

    \item \textbf{Spectral reliability of covariance preconditioning.}
    We empirically examine the relationship between covariance and Hessian eigenspaces, and observe that top covariance components are substantially more aligned with the Hessian than the lower-eigenvalue components.

    \item \textbf{Subspace-aware Kronecker optimizer.}
    Motivated by the analysis above, we propose an optimizer that applies eigenvalue-based preconditioning in the top Kronecker eigenspace and uses an adaptive isotropic acceleration constant in the complementary bottom subspace.
\end{itemize}

\section{Preliminaries}

\subsection{Notations}

Let $\Theta\in\mathbb{R}^{m\times n}$ be a matrix-valued parameter and $G=\nabla_\Theta \ell(\Theta)\in\mathbb{R}^{m\times n}$ its stochastic gradient. We write $g=\mathrm{vec}(G)\in\mathbb{R}^{mn}$ and $C:=\mathbb{E}[gg^\top]\in\mathbb{S}_{++}^{mn}$ for the gradient covariance matrix, where $\mathbb{S}_{++}^{mn}$ denotes the cone of $mn\times mn$ symmetric positive-definite (SPD) matrices.\footnote{Strictly speaking, $C$ is generally positive semidefinite. Following \cite{lin2026understanding}, we consider the positive-definite setting, which can be realized by damping $C+\kappa I$ or by using a sufficiently large batch.}. We denote the Hessian by $H:=\nabla_\theta^2\mathcal{L}(\theta)\in\mathbb{R}^{mn\times mn}$, where $\theta=\mathrm{vec}(\Theta)$. Structured preconditioning approximates $C$ by a Kronecker product $C\approx L\otimes R$, where $L\in\mathbb{S}_{++}^{m}$ and $R\in\mathbb{S}_{++}^{n}$. The corresponding preconditioned gradient satisfies $(L\otimes R)^{-1/2}g \Longleftrightarrow L^{-1/2}GR^{-1/2}$. For a square matrix $A$, $\mathrm{Tr}(A)$ and $\det(A)$ denote its trace and determinant. For matrices of the same size, $A\odot B$ denotes the Hadamard product, and $\|A\|_F$ denotes the Frobenius norm. We use standard matrix differential notation: $DF(S)[H]$ denotes the directional derivative of $F$ at $S$ along $H$, and $\nabla^2F(S)[H]$ denotes the Hessian operator.

\subsection{The original Shampoo algorithm}

Shampoo~\cite{gupta2018shampoo} is a matrix-structured optimizer that maintains two marginal second-moment factors, $L=\mathbb{E}[GG^\top]$ and $R=\mathbb{E}[G^\top G]$. In practice, they are updated by EMA: $L_t=\beta_2L_{t-1}+(1-\beta_2)G_tG_t^\top$ and $R_t=\beta_2R_{t-1}+(1-\beta_2)G_t^\top G_t$. The update is $\Theta_{t+1}=\Theta_t-\eta L_t^{-p}G_tR_t^{-p}$, with common choices $p=1/4$ or $p=1/2$. Thus, Shampoo replaces a dense $mn\times mn$ preconditioner with two smaller factors of sizes $m\times m$ and $n\times n$.

\subsection{Bregman matrix divergences}
Following~\cite{lin2026understanding}, we measure the discrepancy between $C$ and $L\otimes R$ using Bregman matrix divergences. Let $F:\mathbb{S}_{++}^{d}\to\mathbb{R}$ be a strictly convex differentiable generating function. For $X,Y\in\mathbb{S}_{++}^{d}$,
\vspace{-0.6em}
\begin{equation}
\mathcal{B}_F(X,Y)
:=
F(X)-F(Y)-\mathrm{Tr}\big([\nabla F(Y)](X-Y)\big).
\label{eq:bregman_div}
\end{equation}
Different choices of $F$ induce different notions of covariance approximation.

\section{Rethinking Bregman Divergences in Kronecker-Factored Approximation}
\label{sec:rethink_bregman}

Lin et al.~\cite{lin2026understanding} introduced a Bregman-divergence-based view of Kronecker-factored approximation to the gradient covariance matrix, providing a unified interpretation of several structured preconditioners. Specifically, given the covariance matrix $C=\mathbb{E}[gg^\top]$, one seeks a Kronecker-factored SPD approximation $S=L\otimes R$ by solving $\min_{L,R}\mathcal{B}_F(C,L\otimes R)$.

Different generating functions $F$ induce different stationary conditions, and thus derive different bilateral preconditioners. Table~\ref{tab:bregman_precond} summarizes the main cases considered in this paper.

\begin{table}[htbp]
\centering
\caption{Bregman-divergence view of bilateral Kronecker preconditioners.}
\label{tab:bregman_precond}
\resizebox{\textwidth}{!}{
\begin{tabular}{llll}
\toprule
\textbf{Method} & \textbf{$F(M)$} & \textbf{Divergence} & \textbf{Stationary conditions} \\
\midrule
KL-Shampoo 
& $-\frac{1}{2}\log\det M$ 
& LogDet / KL 
& $L^*=\frac{1}{n}\mathbb{E}[G(R^*)^{-1}G^\top]$, 
$R^*=\frac{1}{m}\mathbb{E}[G^\top(L^*)^{-1}G]$ \\[1.2ex]

VN-Shampoo 
& $\mathrm{Tr}(M\log M-M)$ 
& von Neumann 
& $L^*=\frac{1}{\mathrm{Tr}(R^*)}\mathbb{E}[GG^\top]$, 
$R^*=\frac{1}{\mathrm{Tr}(L^*)}\mathbb{E}[G^\top G]$ \\[1.2ex]

F-Shampoo 
& $\frac{1}{2}\mathrm{Tr}(M^\top M)$ 
& Frobenius 
& $L^*=\frac{1}{\mathrm{Tr}((R^*)^2)}\mathbb{E}[GR^*G^\top]$, 
$R^*=\frac{1}{\mathrm{Tr}((L^*)^2)}\mathbb{E}[G^\top L^*G]$ \\
\bottomrule
\end{tabular}
}
\end{table}
Next, we will interpret the different bilateral preconditioners obtained with different divergences from three new perspectives.
\subsection{Unavoidable Approximation Error}
While these optimizers admit a unified Bregman-divergence view, approximating a dense covariance matrix by two low-dimensional factors generally cannot achieve zero divergence, and hence cannot yield perfect spectral matching, which is formalized by the following proposition.

\begin{proposition}[Strict Positivity of Kronecker Approximation Error]
\label{prop:positive_error}
Let $C \in \mathbb{S}_{++}^{mn}$ be a SPD matrix drawn from a distribution that is absolutely continuous with respect to the Lebesgue measure on $\mathbb S^{mn}$. Let $\mathcal{M} = \{L \otimes R \mid L \in \mathbb{S}_{++}^m, R \in \mathbb{S}_{++}^n\}$ denote the manifold of Kronecker-factored matrices. Then, $C \notin \mathcal{M}$ almost surely (a.s.). Consequently, for any Bregman matrix divergence $\mathcal{B}_F$, the minimum approximation error remains strictly positive: $\inf_{L, R} \mathcal{B}_F(C, L \otimes R) > 0 \quad \text{a.s.}$.
\end{proposition}

Given that these Kronecker products cannot perfectly approximate the covariance matrix $\mathbb{E}[gg^\top]$, how do these divergences actually differ in matrix approximation? The following subsection shows the spectral preferences of different divergences.

\subsection{Spectral Preferences of Matrix Divergences}
\label{sec:spec_pref}

According to Proposition~\ref{prop:positive_error}, exact spectrum matching is impossible for a general dense covariance matrix, since $C\notin\mathcal M$. Therefore, different divergences differ not only in the value of the approximation error, but also in how this error is distributed across the spectrum.

To characterize this effect, let $A=U\Lambda U^\top,$ $B=V\Omega V^\top$ with eigenvalues $\{\lambda_i\}_{i=1}^d$ and $\{\omega_j\}_{j=1}^d$, and define the eigenspace alignment matrix $P_{ij}:=\langle u_i,v_j\rangle^2.$ For any spectral generating function $F$, the corresponding Bregman divergence admits the decomposition
\vspace{-0.3em}
\begin{align}
\mathcal{B}_F(A,B)=\Psi(\Lambda)+\Phi(\Omega)+\sum_{i,j} g_F(\lambda_i,\omega_j)P_{ij},
\label{eq:cross_spectral_decomp}
\end{align}
where the first two terms depend only on marginal spectra, while $g_F$ controls the penalty on cross-spectral mismatch. The proof is given in Appendix~\ref{sec:proof_spec_pref}. For the divergences studied in this paper, $g_{\mathrm{Frob}}=-\lambda_i\omega_j$, $g_{\mathrm{vN}}=-\lambda_i\log\omega_j$, and $g_{\mathrm{LogDet}}=\lambda_i/\omega_j$. Thus, Frobenius emphasizes large-eigenvalue components, von Neumann retains this preference but weakens it logarithmically, and LogDet is more sensitive to small-eigenvalue components. Table~\ref{tab:spectral_bias} summarizes the differences.
\begin{table}[htbp]
\centering
\caption{Spectral preferences of matrix divergences.}
\vspace{-0.6em}
\label{tab:spectral_bias}
\resizebox{\textwidth}{!}{
\begin{tabular}{llll}
\toprule
\textbf{Divergence} & \textbf{Coupling term $g_F$} & \textbf{Error sensitivity} & \textbf{Approximation priority} \\
\midrule
Frobenius 
& $-\lambda_i\omega_j$ 
& absolute, bilinear magnitude 
& strongest emphasis on top spectrum \\

von Neumann 
& $-\lambda_i\log\omega_j$ 
& log-damped magnitude
& soft preference for top spectrum \\

LogDet 
& $\lambda_i/\omega_j$ 
& relative ratio distortion 
& broad spectrum, including small-eigenvalue directions \\
\bottomrule
\end{tabular}
}
\end{table}
\vspace{-1.0em}

\subsection{Stationary Conditions Align with the Spectral Preferences}
In this subsection, we show that the bilateral optimizers induced by the stationary conditions of different divergences align with the spectral preferences in Section~\ref{sec:spec_pref}.
\begin{theorem}[Divergence-induced geometric bias on the Kronecker manifold]
\label{thm:divergence_induced_bias}
Let $C$ be a target SPD matrix, and let $\mathcal{M}$ denote the manifold of Kronecker-factored SPD matrices. Suppose $(L^*,R^*)$ is a stationary point of $\min_{L, R} \mathcal{B}_F(C, L\otimes R)$, and let $S^* := L^*\otimes R^* \in \mathcal{M}$. Then, for every feasible first-order perturbation in the tangent space of $\mathcal{M}$ at $S^*$, i.e., $H\in T_{S^*}\mathcal{M}$,
\vspace{-0.6em}
\begin{align}
\left. D_S \mathcal{B}_F(C,S)[H] \right|_{S=S^*}
=
\left\langle \nabla^2 F(S^*)[S^*-C],\, H \right\rangle
=0,
\end{align}
where $\langle A,B\rangle := \mathrm{Tr}(A^\top B)$. Consequently, we have $\nabla^2 F(S^*)[S^*-C]\perp T_{S^*}\mathcal{M}$.
\end{theorem}
Theorem~\ref{thm:divergence_induced_bias} shows that a stationary Kronecker approximation is determined not by the raw residual $S^*-C$, but by the divergence-weighted residual $\nabla^2F(S^*)[S^*-C]$. Hence, different generating functions generate different local error geometries, and the resulting Kronecker factors align with the corresponding spectral preferences given in Section~\ref{sec:spec_pref}, as stated in the following corollary.

\begin{corollary}
\label{cor:stationary_spectral_bias}
Consider the problem of minimizing $\mathcal{B}_F(C,L\otimes R)$ over Kronecker factors $L$ and $R$, where $C\in\mathbb S_{++}^{mn}$. Let $(L^*,R^*)$ be a stationary point and set $S^*=L^*\otimes R^*$. If the divergence generated by $F$ induces a specific spectral weighting on the approximation error, then $S^*$ is stationary with respect to that weighted geometry. Consequently, different choices of $F$ yield stationary Kronecker factors governed by different weighted geometries, leading to different spectral preferences.
\end{corollary}
This is reflected directly in the stationary conditions. Frobenius yields bilinear terms such as $GR^*G^\top$ and $G^\top L^*G$, which emphasize dominant subspaces. LogDet yields inverse-weighted terms such as $G(R^*)^{-1}G^\top$ and $G^\top(L^*)^{-1}G$, which increase sensitivity to small eigenvalues. The von Neumann case is intermediate: it favors dominant directions through trace-normalized moment matching, but avoids the inverse weighting of LogDet. Thus, the stationary equations directly show the spectral preferences of the underlying divergences.

\section{On the Covariance and Hessian Matrix}
\label{sec:covaraince_and_hessian}
The empirical Fisher, gradient covariance, true Hessian, and exact FIM are generally different~\cite{gur2018gradient,kunstner2019limitations,thomas2020interplay}. However, prior work shows that the top eigenspaces of the gradient covariance and Hessian can still exhibit strong alignment~\cite{zhu2019anisotropic,wu2022alignment,zhu2026accelerating}. This suggests that the top spectrum of covariance may contain useful curvature information, even when the full spectrum is unreliable.

\par We verify this by measuring Hessian--covariance eigenspace alignment on a small Transformer trained on SST-2-1k~\cite{song2025does}. The results are reported in Table~\ref{tab:hess_cov_alignment}; details are given in Appendix~\ref{app:hessian_cov_alignment}.
\begin{table}[htbp]
\vspace{-0.6em}
\centering
\caption{Hessian--covariance eigenspace alignment on Q/K/V projection blocks.}
\label{tab:hess_cov_alignment}
\vspace{-0.7em}
\footnotesize
\setlength{\tabcolsep}{3.2pt}
\renewcommand{\arraystretch}{0.92}
\begin{tabular}{c|ccc|ccc}
\toprule
\multirow{2}{*}{Metric}
& \multicolumn{3}{c|}{Step 100}
& \multicolumn{3}{c}{Step 1000} \\
\cmidrule(lr){2-4}\cmidrule(lr){5-7}
& Q & K & V & Q & K & V \\
\midrule
$\mathrm{Overlap@5}$ 
& 0.825 & 0.763 & 0.896 
& 0.804 & 0.748 & 0.855 \\
$\mathrm{BandOverlap@180\text{-}200}$ 
& 0.091 & 0.102 & 0.098 
& 0.097 & 0.105 & 0.103 \\
\bottomrule
\end{tabular}
\vspace{-1.0em}
\end{table}

\section{Combining Bregman Divergence and Covariance Reliability: New Optimizers}
\label{sec:algo}

Sections~\ref{sec:rethink_bregman} and~\ref{sec:covaraince_and_hessian} highlight two observations. First, Kronecker factors cannot perfectly match a dense covariance spectrum, with residuals distributed differently across Bregman divergences. Second, the top covariance eigenspace aligns better with the Hessian, making bottom components unreliable. Prior work also showed that full-covariance preconditioning can perform poorly~\cite{kunstner2019limitations,thomas2020interplay}. Motivated by these observations, we propose a subspace-aware Kronecker preconditioner. The method chooses Kronecker factors via a top-sensitive divergence, such as the von Neumann or Frobenius divergence, applies eigenvalue-based scaling only in the top eigenspace and employs an adaptive isotropic acceleration constant in the bottom subspace. This design uses the reliable top spectral information while avoiding using the noisy bottom space directions.

The full procedure of \textbf{BregTop} is given in Algorithms~\ref{alg:unified_precond} and~\ref{alg:kfac_precond}. Algorithm~\ref{alg:unified_precond} maintains divergence-specific Kronecker factors and their eigenspaces, where \textsc{Stats} returns the factor updates $(\Delta_L,\Delta_R)$ induced by the chosen Bregman divergence. Algorithm~\ref{alg:kfac_precond} then applies the proposed subspace preconditioning rule. The explicit forms of $(\Delta_L,\Delta_R)$ for different divergences are given in Appendix~\ref{app:algo}.
\begin{figure}[htbp]
\centering
\begin{minipage}[t]{0.48\linewidth}
\footnotesize
\captionsetup{type=algorithm,skip=1pt,labelfont=bf}
\hrule
\vspace{2pt}
\caption{Unified subspace-aware optimizer}
\label{alg:unified_precond}
\vspace{2pt}
\hrule
\vspace{2pt}
\begin{algorithmic}[1]
\REQUIRE $G,M,\gamma,\beta_1,\beta_2,T,\rho,q,c$
\STATE $M \leftarrow (1-\beta_1)G+\beta_1 M$
\STATE $(\Delta_L,\Delta_R)\leftarrow \textsc{Stats}(G,U_L,U_R,\lambda_L,\lambda_R)$
\STATE $L \leftarrow (1-\beta_2)\Delta_L+\beta_2 L$, \quad $R \leftarrow (1-\beta_2)\Delta_R+\beta_2 R$\vspace{-12pt}
\STATE $\lambda_L \leftarrow (1-\beta_2)\mathrm{diag}(U_L^\top \Delta_L U_L)+\beta_2\lambda_L$ 
\STATE $\lambda_R \leftarrow (1-\beta_2)\mathrm{diag}(U_R^\top \Delta_R U_R)+\beta_2 \lambda_R$
\IF{iter $\bmod\ T = 0$}
    \STATE $U_L \leftarrow \mathrm{qr}(L U_L)$, \quad $U_R \leftarrow \mathrm{qr}(R U_R)$
\ENDIF
\STATE $\widehat M \leftarrow \textsc{KronPrecond}(U_L,U_R,\lambda_L,\lambda_R,M,\rho,q, c)$
\STATE $\theta \leftarrow \theta-\gamma\,\mathrm{vec}(\widehat M)$
\end{algorithmic}
\vspace{2pt}
\hrule
\end{minipage}
\hfill
\begin{minipage}[t]{0.48\linewidth}
\footnotesize
\captionsetup{type=algorithm,skip=1pt,labelfont=bf}
\hrule
\vspace{2pt}
\caption{Kronecker-based preconditioning}
\label{alg:kfac_precond}
\vspace{2pt}
\hrule
\vspace{2pt}
\begin{algorithmic}[1]
\REQUIRE $U_L,U_R,\lambda_L,\lambda_R,M,\rho,q,c$
\STATE $\widetilde M \leftarrow U_L^\top M U_R$, $S \leftarrow \lambda_L\lambda_R^\top$, $K \leftarrow \lceil \rho mn\rceil$
\STATE $\Omega_\rho \leftarrow$ indices of the top-$K$ entries of $S$
\STATE $\mathcal B_\rho \leftarrow \{(i,j):(i,j)\notin\Omega_\rho\}$
\STATE $\chi_{\rho,q}\leftarrow c\cdot\mathrm{Quantile}_q\{S_{ij}^{-1/2}:(i,j)\in\mathcal B_\rho\}$
\FOR{each $(i,j)$}
    \STATE $W_{ij}\leftarrow
    \begin{cases}
    S_{ij}^{-1/2}, & (i,j)\in\Omega_\rho,\\
    \chi_{\rho,q}, & (i,j)\in\mathcal B_\rho
    \end{cases}$
\ENDFOR
\STATE $\widetilde M_{\mathrm{pre}} \leftarrow \widetilde M \odot W$, $\widehat M \leftarrow U_L \widetilde M_{\mathrm{pre}} U_R^\top$
\RETURN $\widehat M$
\end{algorithmic}
\vspace{2pt}
\hrule
\end{minipage}
\end{figure}

In Algorithm~\ref{alg:kfac_precond}, the retained set $\Omega_\rho$ contains the largest $\lceil \rho mn\rceil$ entries of the joint Kronecker spectrum $S=\lambda_L\lambda_R^\top$. The complementary set $\mathcal B_\rho$ is treated isotropically using the adaptive constant $\chi_{\rho,q}$, chosen as $c$ times the $q$-quantile of $\{S_{ij}^{-1/2}:(i,j)\in\mathcal B_\rho\}$. Thus, the top subspace uses eigenvalue-based scaling, while the bottom subspace uses a uniform accelerating scale.

The following theorem formalizes the decomposition implemented by Algorithm~\ref{alg:kfac_precond}: it applies eigenvalue-based damping in the top subspace and a uniform acceleration in the bottom subspace.

\begin{theorem}[Equivalent decomposition of the preconditioner]
\label{th:pre_equivalence}
Let $L=U_L\Lambda_LU_L^\top$ and $R=U_R\Lambda_RU_R^\top$, and define $S=\lambda_L\lambda_R^\top$. For a retained ratio $\rho\in(0,1]$, let $\Omega_\rho$ contain the indices of the largest $K=\lceil \rho mn\rceil$ entries of $S$, and let $E$ be its indicator matrix. Define $\mathcal P_{\mathrm{top}}^\rho(M):=U_L\big((U_L^\top M U_R)\odot E\big)U_R^\top$. Then Algorithm~\ref{alg:kfac_precond} returns
\begin{align}
\widehat M=U_L\Big((U_L^\top M U_R)\odot E \odot S^{-1/2}\Big)U_R^\top+\chi_{\rho,q}\big(M-\mathcal P_{\mathrm{top}}^\rho(M)\big).
\end{align}
Thus, the top Kronecker eigenspace is scaled by its eigenvalue-based inverse curvature, while the orthogonal bottom space is scaled by the adaptive acceleration constant $\chi_{\rho,q}$ given in Algorithm~\ref{alg:kfac_precond}.
\end{theorem}

\begin{remark}
While Song et al.~\cite{song2025does} showed that learning primarily occurs in the Hessian's non-dominant (bottom) subspace, our work studies the spectral alignment between the Hessian and its Kronecker covariance approximation. The optimizer design is consistent with the training dynamics observed in \cite{song2025does}: preconditioning in the top subspace dampens noisy updates along high-curvature directions, while isotropic acceleration in the bottom subspace preserves the standard gradient flow, which is necessary for the learning progress.
\end{remark}

\section{Experiments}

We compare Shampoo, KL-Shampoo, VN-Shampoo, F-Shampoo, and BregTop on full SST-2. The model is a 4-layer Transformer with hidden dimension 128 and 8 attention heads. For all methods, Adam is used for the embedding, bias, normalization, and classification-head parameters, while the Shampoo-type optimizer is applied to the remaining matrix weights. More details are given in Appendix~\ref{app:sst2_optimization_details}. BregTop-VN and BregTop-F use the algorithm in Section~\ref{sec:algo} with VN/F-induced Kronecker factors, respectively. We report the number of steps required for the EMA-smoothed training loss to fall below $0.05$ with EMA $0.98$. No weight decay is used.

\begin{table}[htbp]
\centering
\caption{Steps to EMA-smoothed train loss $<0.05$ on full SST-2.}
\label{tab:sst2_pretrain_steps}
\scriptsize
\setlength{\tabcolsep}{2.5pt}
\resizebox{\linewidth}{!}{
\begin{tabular}{lcccccccccc}
\toprule
 & Shampoo & KL & VN-v1 & VN-v2 & F-v1 & F-v2 & B-VN-v1 & B-VN-v2 & B-F-v1 & B-F-v2 \\
\midrule
Steps & 1119 & 1092 & 1141 & 1127 & 1598 & 1602 & 1018 & \textbf{1003} & 1084 & 1059 \\
\bottomrule
\end{tabular}
}
\vspace{-1.0em}
\end{table}

BregTop-VN-v2 reaches the target loss fastest. The BregTop variants improve over their corresponding VN/F baselines, supporting the benefit of our algorithm.

\section{Conclusion}

We studied Bregman-induced Kronecker preconditioning under unavoidable covariance approximation error. Our analysis shows that different divergences impose different spectral preferences, while our empirical results suggest that only the top eigenspace is reliably aligned with the Hessian. Based on this, we proposed a subspace-aware optimizer that trusts the top eigenspace and treats the bottom space with adaptive isotropic scaling. Experiments show that this design improves the step efficiency.

\bibliography{sample}
\newpage
\clearpage
\appendix

\section{Related Work}
\label{sec:related_work}
\paragraph{Matrix-Based Optimizers.}
Element-wise optimizers, such as Adam~\cite{kingma2014adam,loshchilov2018decoupled}, have long dominated large-scale neural network  training, but they ignore the natural matrix structure of gradients. This limitation has motivated a growing line of work on matrix-based optimizers, including bilaterally preconditioned methods such as K-FAC~\cite{martens2015optimizing}, Shampoo~\cite{gupta2018shampoo}, SOAP~\cite{vyas2025soap}, and FISMO~\cite{xu2026fismo}, as well as one-sided optimizers such as Muon~\cite{jordan2024muon,khaled2025muonbp}, ASGO~\cite{an2025asgo}, and SSO~\cite{xie2026controlled}. These methods have shown strong empirical performance and increasing practical relevance in large-scale training.

\paragraph{Anisotropic Training Dynamics.}
A growing body of work has shown that the loss landscape of neural networks is highly anisotropic and ill-conditioned. In particular, the Hessian spectrum typically consists of a small number of large eigenvalues forming a dominant top subspace, while the vast majority of eigenvalues are close to zero and form a low-curvature bulk subspace~\cite{wangsharpness,wenunderstanding}. Although a substantial part of the gradient energy lies in the dominant eigendirections, these components often exhibit strong oscillation and may slow down optimization \cite{gur2018gradient,song2025does}. In contrast, some low-curvature directions can be important for sustained progress, but their curvature estimates are often noisy and require careful stabilization. Motivated by this observation, several recent optimizers have explicitly adopted a top-space damping and tiny-subspace acceleration design to improve training efficiency~\cite{zhou2025bsfa,zhu2026accelerating}.

\paragraph{Understanding Structured Preconditioners.}
While Shampoo~\cite{gupta2018shampoo} and its variants have achieved remarkable empirical success~\cite{kasimbeg2025accelerating}, a recent line of work aims to theoretically explain their effectiveness. In particular, \cite{morwaninew2025,eschenhagen2026purifying,xie2025structured} interpret Shampoo-like preconditioners through the lens of Kronecker-product approximations $L\otimes R$ to the gradient covariance matrix $\mathbb{E}[gg^\top]$, typically evaluated under the Frobenius norm. More recently, Lin et al.~\cite{lin2026understanding} have further generalized this perspective by studying these preconditioners through the framework of \emph{matrix divergences}. This view shows that different bilateral Shampoo variants can be interpreted as minimizing different Bregman divergences between the dense covariance matrix and its Kronecker approximation. It therefore provides a unified language for comparing Frobenius-, von Neumann-, and LogDet/KL-induced preconditioners.

\section{Proofs and Additional Details}
\subsection{Proof of Proposition \ref{prop:positive_error}}
The dimension of the ambient space $\mathbb{S}_{++}^{mn}$ is $\frac{mn(mn+1)}{2} -1= \mathcal{O}(m^2 n^2)$. The Kronecker manifold $\mathcal{M}$ is parameterized by $L$ and $R$, yielding at most $\frac{m(m+1)}{2} + \frac{n(n+1)}{2} = \mathcal{O}(m^2 + n^2)$ degrees of freedom. Because $\text{dim}(\mathcal{M}) \ll \text{dim}(\mathbb{S}_{++}^{mn})$, $\mathcal{M}$ is a strictly lower-dimensional submanifold, which has Lebesgue measure zero in $\mathbb{S}_{++}^{mn}$.

Since $C$ admits a density with respect to the Lebesgue measure, we have $\mathbb P(C\in\mathcal M)=0$, so $C\notin\mathcal M$ almost surely. For the Bregman divergences considered in this paper, $\mathcal B_F(C,S)\to0$ implies $S\to C$. Since the Kronecker-factored SPD family is closed under positive-definite limits, $\inf_{L,R}\mathcal B_F(C,L\otimes R)=0$ would imply $C\in\mathcal M$, which contradicts $C\notin\mathcal M$ almost surely. Therefore, we can derive
\begin{align}
\inf_{L,R}\mathcal B_F(C,L\otimes R)>0
\quad\text{a.s.} .   
\end{align}

\subsection{Proof of Section~\ref{sec:spec_pref}}
\label{sec:proof_spec_pref}
By the definition of the Bregman matrix divergence generated by a strictly convex, differentiable function $F$, we have:
\begin{align}
\mathcal{B}_F(A, B) = F(A) - F(B) - \mathrm{Tr}\big(\nabla F(B)(A - B)\big).
\end{align}
For a spectral generating function $F(M) = \sum_{k} f(\lambda_k(M)) = \mathrm{Tr}(f(M))$, its matrix gradient is computed directly via the scalar derivative applied to its eigenvalues: $\nabla F(M) = f'(M)$. Consequently, given the spectral decomposition $B = V\Omega V^\top$, the gradient can be written as $\nabla F(B) = V f'(\Omega) V^\top$. Using the linearity of the trace, we can expand the cross-term:
\begin{align}
\mathrm{Tr}\big(\nabla F(B)(A - B)\big) = \mathrm{Tr}\big(\nabla F(B) A\big) - \mathrm{Tr}\big(\nabla F(B) B\big).
\end{align}
The second term depends only on the spectrum of $B$:

\begin{align}
\mathrm{Tr}\big(\nabla F(B) B\big) = \mathrm{Tr}\big(V f'(\Omega) V^\top V \Omega V^\top\big) = \mathrm{Tr}\big(f'(\Omega) \Omega\big) = \sum_{j=1}^d f'(\omega_j) \omega_j. 
\end{align}

For the cross-trace term $\mathrm{Tr}(\nabla F(B) A)$, we substitute $A = U\Lambda U^\top$ and apply the cyclic property of the trace:
\begin{align}
\mathrm{Tr}\big(\nabla F(B) A\big) = \mathrm{Tr}\big(V f'(\Omega) V^\top U \Lambda U^\top\big) = \mathrm{Tr}\big(f'(\Omega) V^\top U \Lambda U^\top V\big).
\end{align}
Let $Q = U^\top V$ represent the orthogonal transition matrix between the eigenspaces of $A$ and $B$. By definition, squaring its entries gives the eigenspace alignment matrix $P$, such that $P_{ij} = Q_{ij}^2 = \langle u_i, v_j \rangle^2$. Thus, the cross-trace becomes:
\begin{align}
\mathrm{Tr}\big(\nabla F(B) A\big) = \mathrm{Tr}\big(f'(\Omega) Q^\top \Lambda Q\big) = \sum_{i=1}^d \sum_{j=1}^d f'(\omega_j) \lambda_i Q_{ij}^2 = \sum_{i=1}^d \sum_{j=1}^d f'(\omega_j) \lambda_i P_{ij}.
\end{align}
Substituting these decoupled components back into the Bregman divergence definition, we obtain:
\begin{align}
\mathcal{B}_F(A, B) = F(A) - F(B) + \mathrm{Tr}\big(\nabla F(B) B\big) - \mathrm{Tr}\big(\nabla F(B) A\big).
\end{align}
We can now clearly see the three distinct parts of the decomposition:
\begin{align}
    &\Psi(\Lambda) = F(A) = \sum_{i=1}^d f(\lambda_i),\quad \Phi(\Omega) = \mathrm{Tr}\big(\nabla F(B) B\big) - F(B) = \sum_{j=1}^d \big( f'(\omega_j)\omega_j - f(\omega_j) \big),\\
    &\sum_{i=1}^d \sum_{j=1}^d g_F(\lambda_i, \omega_j) P_{ij} = - \mathrm{Tr}\big(\nabla F(B) A\big) \implies g_F(\lambda_i, \omega_j) = - \lambda_i f'(\omega_j).
\end{align}
For the three divergences considered in this paper, the coupling terms are:

\textbf{Frobenius Divergence:} Let $F(M) = \frac{1}{2}\mathrm{Tr}(M^2) \implies f(x) = \frac{1}{2}x^2$, yielding $f'(x) = x$.
\begin{align}
g_{\text{Frob}}(\lambda_i, \omega_j) = - \lambda_i (\omega_j) = - \lambda_i \omega_j.
\end{align}

\textbf{von Neumann Divergence:} Let $F(M) = \mathrm{Tr}(M \log M - M) \implies f(x) = x \log x - x$, yielding $f'(x) = \log x$.
\begin{align}
g_{\text{vN}}(\lambda_i, \omega_j) = - \lambda_i (\log \omega_j) = - \lambda_i \log \omega_j.
\end{align}

\textbf{Log-Determinant Divergence\footnote{Up to an irrelevant positive scaling constant, we take $F(M)=-\log\det M$ for the LogDet case.}:} Let $F(M) = -\log \det M = -\mathrm{Tr}(\log M) \implies f(x) = -\log x$, yielding $f'(x) = -1/x$.
\begin{align}
g_{\text{LogDet}}(\lambda_i, \omega_j) = - \lambda_i \left(-\frac{1}{\omega_j}\right) = \frac{\lambda_i}{\omega_j}.
\end{align}
This completes the proof.

\begin{remark}
The function $g_F$ serves as the ``exchange rate'' for approximation errors. By analyzing $g_F$, we can derive the spectral bias of each divergence:

\begin{itemize}
    \item \textbf{Absolute-error emphasis (Frobenius and von Neumann).}
    In both $\mathcal{B}_{\text{Frob}}$ and $\mathcal{B}_{\text{vN}}$, the cross-spectral term is weighted by the target eigenvalue $\lambda_i$, either linearly through $\lambda_i \omega_j$ or through $\lambda_i \log \omega_j$. As a result, mismatches of small eigenvalues contribute relatively little to the divergence, while errors on large eigenvalue components are penalized more strongly. Therefore, these divergences are biased toward the top spectrum: when approximation error is unavoidable, they are more likely to fit the dominant eigenspace.

    \item \textbf{Relative-error emphasis (Log-Determinant).}
    For $\mathcal{B}_{\mathrm{LogDet}}$, the coupling term is $\lambda_i/\omega_j$. The penalty depends on relative scale rather than only on absolute magnitude. As a result, small eigenvalue components are not ignored, and the divergence puts more weight on preserving relative spectral accuracy across the whole spectrum.
\end{itemize}
\end{remark}

\subsection{BregTop Algorithmic Details}
\label{app:algo}
The statistics $(\Delta_L,\Delta_R)=\textsc{Stats}(G,U_L,U_R,\lambda_L,\lambda_R)$ in Algorithm~\ref{alg:unified_precond} are given as follows. These updates follow the divergence-induced Kronecker statistics derived in~\cite{lin2026understanding}:
\begin{align}
\left( \Delta_L, \Delta_R \right) := 
\begin{cases} 
\left( GG^\top,\; G^\top G \right) 
& \text{(Original)} \\[6pt]
\left( GG^\top/ \mathrm{Tr}(R),\; G^\top G/ \mathrm{Tr}(L) \right) 
& \text{(VN-v1)} \\[6pt]
\left( GG^\top / \sum\limits_{i=1}^n \lambda_{R,i},\;
G^\top G / \sum\limits_{i=1}^m \lambda_{L,i} \right) 
& \text{(VN-v2)} \\[8pt]
\left( G R G^\top / \mathrm{Tr}(R^2),\;
G^\top L G / \mathrm{Tr}(L^2) \right) 
& \text{(F-v1)} \\[6pt]
\left( G U_R \mathrm{Diag}(\lambda_R) U_R^\top G^\top / \sum\limits_{i=1}^n \lambda_{R,i}^2,\;
G^\top U_L \mathrm{Diag}(\lambda_L) U_L^\top G / \sum\limits_{i=1}^m \lambda_{L,i}^2 \right) 
& \text{(F-v2)}
\end{cases}
\end{align}

\subsection{Proof of Theorem~\ref{thm:divergence_induced_bias}}
Define the reduced objective
\begin{align}
\phi(L,R) := \mathcal{B}_F(C,L\otimes R).
\end{align}
Stationarity of $(L^*,R^*)$ implies that, for any feasible perturbation $(\Delta L,\Delta R)$, the first-order directional derivative of $\phi$ is zero:
\begin{align}
D\phi(L^*,R^*)[\Delta L,\Delta R] = 0
\qquad
\forall\, \Delta L \in \mathbb{S}^{m},\; \Delta R \in \mathbb{S}^{n}.
\end{align}

We first characterize the tangent space of $\mathcal{M}$ at $S^*=L^*\otimes R^*$. Consider the smooth curve
\begin{align}
S(t) := (L^*+t\Delta L)\otimes (R^*+t\Delta R).
\end{align}
Then
\begin{align}
\frac{d}{dt}S(t)\Big|_{t=0}
=
\Delta L \otimes R^* + L^* \otimes \Delta R.
\end{align}
Hence, the tangent space is
\begin{align}
T_{S^*}\mathcal{M}
=
\left\{
\Delta L \otimes R^* + L^* \otimes \Delta R
\;:\;
\Delta L \in \mathbb{S}^{m},\;
\Delta R \in \mathbb{S}^{n}
\right\}.
\end{align}
Next, by the chain rule,
\begin{align}
D\phi(L^*,R^*)[\Delta L,\Delta R]
=
D_S\mathcal{B}_F(C,S^*)
\big[
\Delta L \otimes R^* + L^* \otimes \Delta R
\big].
\end{align}
Since the left-hand side is zero for all $(\Delta L,\Delta R)$, we obtain
\begin{align}
D_S\mathcal{B}_F(C,S^*)[H] = 0
\qquad
\forall\, H \in T_{S^*}\mathcal{M}.
\end{align}
We differentiate the Bregman divergence with respect to its second argument. From the definition,
\begin{align}
\mathcal{B}_F(C,S)
=
F(C)-F(S)-\langle \nabla F(S),\, C-S\rangle.
\end{align}
Fix any direction $H\in\mathbb{S}^{mn}$. Using Fr\'echet differentiation, we derive
\begin{align}
D_S\mathcal{B}_F(C,S)[H]
=
- \langle \nabla F(S), H\rangle
- \left(
\langle \nabla^2F(S)[H],\, C-S\rangle
-
\langle \nabla F(S), H\rangle
\right).
\end{align}
The two first-order terms cancel, yielding
\begin{align}
D_S\mathcal{B}_F(C,S)[H]
=
\langle \nabla^2F(S)[H],\, S-C\rangle.
\end{align}
Using the self-adjointness of $\nabla^2F(S)$ under the Frobenius inner product, we equivalently write
\begin{align}
D_S\mathcal{B}_F(C,S)[H]
=
\langle \nabla^2F(S)[S-C],\, H\rangle.
\end{align}
Evaluating at $S=S^*$ gives
\begin{align}
D_S\mathcal{B}_F(C,S^*)[H]
=
\left\langle
\nabla^2F(S^*)[S^*-C],\, H
\right\rangle.
\end{align}
Combining all the proof above, for every $H \in T_{S^*}\mathcal{M}$ we have
\begin{align}
\left\langle
\nabla^2F(S^*)[S^*-C],\, H
\right\rangle = 0,
\end{align}
which is exactly
\begin{align}
\nabla^2F(S^*)[S^*-C]
\perp
T_{S^*}\mathcal{M}.
\end{align}
Thus the divergence-weighted error is orthogonal to all feasible first-order directions on the Kronecker manifold. This completes the proof.

\subsection{Proof of Theorem~\ref{th:pre_equivalence}}
Let $M \in \mathbb{R}^{m \times n}$ be the input momentum matrix. Let $U_L \in \mathbb{R}^{m \times m}$ and $U_R \in \mathbb{R}^{n \times n}$ be the orthogonal matrices from the eigendecomposition, satisfying $U_L U_L^\top = I_m$ and $U_R U_R^\top = I_n$. Let $S \in \mathbb{R}^{m \times n}$ be the eigenvalue matrix where $S_{i,j} = \lambda^{(L)}_i \lambda^{(R)}_j$.

Algorithm~\ref{alg:kfac_precond} rotates $M$ into the Kronecker eigenspace to obtain $\widetilde{M} = U_L^\top M U_R$. 
We define the top eigenspace set as $\Omega_\rho$, the indices of the largest $K=\lceil \rho mn\rceil$ entries of $S$. Let $E \in \{0,1\}^{m \times n}$ be the indicator matrix for the top space, where $E_{i,j} = 1$ if $(i,j) \in \Omega_\rho$, and $0$ otherwise. Consequently, the indicator matrix for the orthogonal bottom space is $E^\perp = \mathbf{1} - E$, where $\mathbf{1}$ is the matrix of all ones.

By construction, the scaling matrix $W$ decomposes as
\begin{align}
W = (S^{-\frac{1}{2}} \odot E) + (\chi_{\rho,q} \mathbf{1} \odot E^\perp).    
\end{align}
Since $\mathbf{1} \odot E^\perp = E^\perp = \mathbf{1} - E$, we have:
\begin{align}
 W = (S^{-\frac{1}{2}} \odot E) + \chi_{\rho,q} (\mathbf{1} - E).    
\end{align}

The final output of the algorithm $\widehat{M}$, is computed by applying the mask and rotating back:
$$ \widehat{M} = U_L (\widetilde{M} \odot W) U_R^\top $$

Substituting the decomposed $W$ into the equation and utilizing the distributive property of the Hadamard product:
$$ \widehat{M} = U_L \left( \widetilde{M} \odot \left[ (S^{-\frac{1}{2}} \odot E) + \chi_{\rho,q} \mathbf{1} - \chi_{\rho,q} E \right] \right) U_R^\top $$
$$ \widehat{M} = U_L \left( \widetilde{M} \odot (S^{-\frac{1}{2}} \odot E) \right) U_R^\top + U_L (\chi_{\rho,q} \widetilde{M} \odot \mathbf{1}) U_R^\top - U_L (\chi_{\rho,q} \widetilde{M} \odot E) U_R^\top $$

Notice that $\widetilde{M} \odot \mathbf{1} = \widetilde{M}$. The second term can be simplified by the orthogonality of $U_L$ and $U_R$:
$$ U_L (\chi_{\rho,q} \widetilde{M}) U_R^\top = \chi_{\rho,q} U_L (U_L^\top M U_R) U_R^\top = \chi_{\rho,q} (U_L U_L^\top) M (U_R U_R^\top) = \chi_{\rho,q} I_m M I_n = \chi_{\rho,q} M $$

By rearranging the terms, we obtain:
$$ \widehat{M} = U_L \left( \widetilde{M} \odot (S^{-\frac{1}{2}} \odot E) \right) U_R^\top + \chi_{\rho,q} M - \chi_{\rho,q} U_L (\widetilde{M} \odot E) U_R^\top.$$

We now verify that $\mathcal{P}^\rho_{\mathrm{top}}(M) := U_L (\widetilde{M} \odot E) U_R^\top$ is an exact orthogonal projection of $M$ onto the top eigenspace.

Although the retained index set $\Omega_\rho$ typically forms a staircase shape rather than a rectangle, the orthogonal projection property holds for any arbitrary subset. For notational convenience and without loss of generality, we assume $E$ isolates a top-left $b \times c$ block:
\begin{align}
    \widetilde{M} \odot E = \begin{bmatrix} \widetilde{M}_{b \times c} & \mathbf{0} \\ \mathbf{0} & \mathbf{0} \end{bmatrix}.
\end{align}
Write $U_L=[\widetilde U_L\; U_{L\perp}]$ and $U_R=[\widetilde U_R\; U_{R\perp}]$, where $\widetilde U_L\in\mathbb R^{m\times b}$ and $\widetilde U_R\in\mathbb R^{n\times c}$ contain the retained eigenvectors. Then block multiplication gives
\begin{align}
U_L \begin{bmatrix} \widetilde{M}_{b \times c} & \mathbf{0} \\ \mathbf{0} & \mathbf{0} \end{bmatrix} U_R^\top &= [ \widetilde{U}_L \quad U_{L\perp} ] \begin{bmatrix} \widetilde{M}_{b \times c} & \mathbf{0} \\ \mathbf{0} & \mathbf{0} \end{bmatrix} \begin{bmatrix} \widetilde{U}_R \quad U_{R\perp} \end{bmatrix}^\top \\
&= [\widetilde{U}_L \widetilde{M}_{b \times c}\quad \mathbf{0}] \begin{bmatrix} \widetilde{U}_R \quad U_{R\perp} \end{bmatrix}^\top = \widetilde{U}_L \widetilde{M}_{b \times c} \widetilde{U}_R^\top.
\end{align}
Substituting the projection $\widetilde{M}_{b \times c} = \widetilde{U}_L^\top M \widetilde{U}_R$ into the expression yields:
\begin{align}
\mathcal{P}^\rho_{\mathrm{top}}(M) = (\widetilde{U}_L \widetilde{U}_L^\top) M (\widetilde{U}_R \widetilde{U}_R^\top) = P_L M P_R,
\end{align}
where $P_L = \widetilde{U}_L \widetilde{U}_L^\top$ and $P_R = \widetilde{U}_R \widetilde{U}_R^\top$ are the orthogonal projectors onto the top subspaces of $L$ and $R$, respectively. By the properties of the Kronecker product, this is equivalent to:
\begin{align}
\text{vec}(\mathcal{P}^\rho_{\mathrm{top}}(M)) = (P_R \otimes P_L) \text{vec}(M).
\end{align} 
Since the Kronecker product of two orthogonal projectors is itself an orthogonal projector, this confirms that $U_L (\widetilde{M} \odot E) U_R^\top$ is the exact orthogonal projection of $M$ onto the subspace spanned by the top $bc$ eigenvectors of $L \otimes R$. Consequently, the term $U_L \left( (\widetilde{M} \odot E) \odot S^{-\frac{1}{2}} \right) U_R^\top$ represents the preconditioned momentum where each eigen-component within this top subspace is scaled by the inverse square root of its corresponding eigenvalue.

By definition, the projection onto the orthogonal bottom space is $\mathcal{P}_{\text{bot}}(M) = M - \mathcal{P}^\rho_{\mathrm{top}}(M)$.
\begin{align}
\widehat{M} = \underbrace{U_L \left( (\widetilde{M} \odot E) \odot S^{-\frac{1}{2}} \right) U_R^\top}_{\text{Precise eigenvalue damping on top space}} + \underbrace{\chi_{\rho,q} \left( M - \mathcal{P}^\rho_{\mathrm{top}}(M) \right)}_{\text{Uniform acceleration on bottom space}}
\end{align}

The expression explicitly demonstrates that the global masking operation is mathematically identical to projecting the gradient onto the top eigenspace for precise damping, while independently applying an adaptive uniform acceleration $\chi_{\rho,q}$ to the remaining orthogonal bottom space.

\section{Details of Hessian--Covariance Alignment Experiments}
\label{app:hessian_cov_alignment}

We follow the same setup used in prior Hessian studies \cite{song2025does,damian2023selfstabilization}. The model is a two-layer Transformer with hidden dimension $64$ and $8$ attention heads, trained with SGD on the first $1000$ samples of SST-2 dataset using the MSE loss. For each checkpoint, we evaluate the Q/K/V matrices of the second Transformer layer.

For each selected block, we collect $N=200$ mini-batch gradients and form the empirical second-moment matrix $C=N^{-1}\sum_{b=1}^N g_b g_b^\top$, where $g_b$ is the batch-averaged gradient vector of the corresponding Q/K/V block. We compute the top covariance eigenspace from the SVD of the stacked gradient matrix, without explicitly forming $C$.

The Hessian block is computed with respect to the same Q/K/V parameter block. We use Hessian-vector products and the Lanczos routine to obtain the top Hessian eigenvectors. The Hessian loss is evaluated on the same SST-2 subset. We report the top-space alignment
\begin{align}
\left(\mathrm{Overlap@5}=\|(U_H^{1:5})^\top U_C^{1:5}\|_F^2/5\right)^{1/2},
\end{align}
where $U_H^{1:5}$ and $U_C^{1:5}$ denote the top-5 Hessian and covariance eigenspaces, respectively. To probe lower-spectrum covariance directions, we also compute the spectral-band overlap
\begin{align}
\left(\|(U_H^{180:200})^\top U_C^{180:200}\|_F^2/21\right)^{1/2}.
\end{align}

\section{Additional Details for the SST-2 Optimization Experiment}
\label{app:sst2_optimization_details}

\paragraph{Task and model.}
We conduct the optimization-speed comparison on the \textbf{full} SST-2 training set. The model is a randomly initialized Transformer with 4 layers, hidden dimension 128, and 8 attention heads, containing approximately 4.7M parameters. The classifier is trained with mean-squared error loss on one-hot class targets, following \cite{song2025does,damian2023selfstabilization}.

\paragraph{Parameter grouping.}
For all Shampoo-type methods, we use the same parameter grouping. The embedding layer, bias terms, normalization parameters, and classification head are optimized by Adam. The reamaining matrix weights in the are optimized by the corresponding Shampoo-type optimizer. No weight decay is used.

\paragraph{Selected hyperparameters.}
All methods use batch size 128, learning rate $10^{-3}$ for both the matrix optimizer and the Adam group. For BregTop-VN, we use $\rho=0.06$, $c=3.0$, and $q=0.01$. For BregTop-F, we use $\rho=0.05$, $c=1.6$, and $q=0.01$. These values are used for both the v1 and v2 variants of the corresponding BregTop optimizer. For the remaining hyperparameters, such as $\beta_1$ and $\beta_2$, we follow \cite{lin2026understanding}.

\paragraph{Evaluation metric.}
We compare optimization speed by measuring the number of steps required for the EMA-smoothed training loss to fall below $0.05$. The EMA coefficient is set to $0.98$.

\end{document}